\documentclass[11pt,a4paper]{article}
\usepackage[hyperref]{acl2021}
\usepackage{times}
\usepackage{latexsym}

\usepackage{natbib}
\usepackage{graphicx}
\usepackage{booktabs}
\usepackage{float}
\usepackage{multirow}
\usepackage{amsmath,amssymb, amsthm, amstext}
\usepackage{extarrows}
\usepackage{caption}
\usepackage{subcaption}
\usepackage{mathrsfs}
\usepackage{tikz}
\usetikzlibrary{bayesnet}
\usepackage{hyperref}
\usepackage{cleveref}
\usepackage{lipsum}
\usepackage{arydshln}
\usepackage{enumitem}
\usepackage{algorithm}
\usepackage[noend]{algpseudocode}
\usepackage{rotating}
\usepackage{tikz}
\usepackage[normalem]{ulem}
\usepackage{microtype}

\makeatletter
  \newcommand\reduline{\bgroup\markoverwith{\textcolor{red}{\rule[-0.5ex]{2pt}{0.4pt}}}\ULon}
  \UL@protected\def\blueuwave{\leavevmode \bgroup 
    \ifdim \ULdepth=\maxdimen \ULdepth 3.5\p@
    \else \advance\ULdepth2\p@ 
    \fi \markoverwith{\lower\ULdepth\hbox{\textcolor{blue}{\sixly \char58}}}\ULon}
      \UL@protected\def\reduwave{\leavevmode \bgroup 
    \ifdim \ULdepth=\maxdimen \ULdepth 3.5\p@
    \else \advance\ULdepth2\p@ 
    \fi \markoverwith{\lower\ULdepth\hbox{\textcolor{red}{\sixly \char58}}}\ULon}
          \UL@protected\def\blackuwave{\leavevmode \bgroup 
    \ifdim \ULdepth=\maxdimen \ULdepth 3.5\p@
    \else \advance\ULdepth2\p@ 
    \fi \markoverwith{\lower\ULdepth\hbox{\textcolor{black}{\sixly \char58}}}\ULon}
  \UL@protected\def\yellowdotuline{\leavevmode \bgroup 
    \UL@setULdepth
    \ifx\UL@on\UL@onin \advance\ULdepth2\p@\fi
    \markoverwith{\begingroup
       \lower\ULdepth\hbox{\kern.06em \textcolor{yellow}{.}\kern.04em}%
       \endgroup}%
    \ULon}
  \UL@protected\def\greendashuline{\leavevmode \bgroup 
    \UL@setULdepth
    \ifx\UL@on\UL@onin \advance\ULdepth2\p@\fi
    \markoverwith{\kern.13em
    \vtop{\color{green}\kern\ULdepth \hrule width.3em}%
    \kern.13em}\ULon}
\makeatother

\newcommand{\solidbox}[1]{%
\tikz[baseline]\node[anchor=base, draw=black, outer sep=0pt, inner sep=1pt] {#1};%
}
\newcommand{\dashedbox}[1]{%
\tikz[baseline]\node[anchor=base, draw=black, dash pattern=on 2pt off 1pt, outer sep=0pt, inner sep=1pt] {#1};%
}

\makeatletter
\def\adl@drawiv#1#2#3{%
        \hskip.5\tabcolsep
        \xleaders#3{#2.5\@tempdimb #1{1}#2.5\@tempdimb}%
                #2\z@ plus1fil minus1fil\relax
        \hskip.5\tabcolsep}
\newcommand{\cdashlinelr}[1]{%
  \noalign{\vskip\aboverulesep
           \global\let\@dashdrawstore\adl@draw
           \global\let\adl@draw\adl@drawiv}
  \cdashline{#1}
  \noalign{\global\let\adl@draw\@dashdrawstore
           \vskip\belowrulesep}}
\makeatother

\definecolor{lan}{RGB}{0, 153, 51}
\definecolor{ehsan}{RGB}{200, 100, 0}

\aclfinalcopy 

\title{Unsupervised Representation Disentanglement of Text: \\
An Evaluation on Synthetic Datasets}

\author{
  Lan Zhang$^\spadesuit$\ \ \ \ 
  Victor Prokhorov$^\clubsuit$\ \ \ \ 
  Ehsan Shareghi$^\spadesuit$$^\clubsuit$\\
  $^\spadesuit$~Department of Data Science \& AI, Monash University \\
  $^\clubsuit$~Language Technology Lab, University of Cambridge\\
  {\tt lan.zhang@monash.edu}\ \ \ \   {\tt vp361@cam.ac.uk}\\{\tt ehsan.shareghi@monash.edu}
}
\date{}

\begin{document}
\maketitle
\begin{abstract}
To highlight the challenges of achieving representation disentanglement for text domain in an unsupervised setting, in this paper we select a representative set of successfully applied models from the image domain. We evaluate these models on 6 disentanglement metrics, as well as on downstream classification tasks and homotopy. To facilitate the evaluation, we propose two synthetic datasets with known generative factors. Our experiments highlight the existing gap in the text domain and illustrate that certain elements such as representation sparsity (as an inductive bias), or representation coupling with the decoder could impact disentanglement. To the best of our knowledge, our work is the first attempt on the intersection of unsupervised representation disentanglement and text, and provides the experimental framework and datasets for examining future developments in this direction.\footnote{Code and datasets are available at \url{https://github.com/lanzhang128/disentanglement}}

\end{abstract}

\section{Introduction}
Learning task-agnostic unsupervised representations of data has been the center of attention across various areas of Machine Learning and more specifically NLP. However, little is known about the way these continuous representations organise information about data. In recent years, the NLP community has focused on the question of design and selection of suitable linguistic tasks to probe the presence of syntactic or semantic phenomena in representations as a whole~\citep{bosc-vincent-2020-sequence,voita-titov-2020-information,torroba-hennigen-etal-2020-intrinsic,pimentel-etal-2020-information,hewitt-liang-2019-designing,ettinger-etal-2018-assessing,marvin-linzen-2018-targeted,conneau-etal-2018-cram}.
Nonetheless, a fine-grain understanding of information organisation in coordinates of a continuous representation is yet to be achieved. 

Arguably, a necessity to move in this direction is agreeing on the cognitive process behind language generation (fusing semantic, syntactic, and lexical components), which can then be reflected in the design of representation learning frameworks. However, this still remains generally as an area of debate and perhaps less pertinent in the era of self-supervised masked language models and the resulting surge of new state-of-the-art results.
%

Even in the presence of such an agreement, learning to disentangle the surface realization of the underlying factors of data (e.g., semantics, syntactic, lexical) in the representation space is a non-trivial task. Additionally, there is no established study for evaluating such models in NLP. A handful of recent works have looked into disentanglement for text by splitting the representation space into \emph{predefined} disentangled subspaces such as style and content~\cite{cheng-etal-2020-improving,john-etal-2019-disentangled}, or syntax and semantics \cite{balasubramanian-etal-2021-polarized,bao-etal-2019-generating,chen-etal-2019-multi-task}, and rely on \emph{supervision} during training. However, a generalizable and realistic approach needs to be \emph{unsupervised} and capable of identifying the underlying factors solely via the regularities presented in data. 





%


In areas such as image processing, the same question has been receiving a lot of attention and inspired a wave of methods for learning and evaluating unsupervised representation disentanglement~\citep{DBLP:journals/corr/abs-2102-05185,pmlr-v97-mathieu19a,pmlr-v80-kim18b,burgess2018,higgins2018definition,higgins2016} and creation of large scale datasets~\cite{dittadi2021on}. It has been argued that disentanglement is the means towards representation interpretability~\cite{pmlr-v97-mathieu19a}, generalization~\cite{montero2021the}, and robustness~\cite{Bengio2013,DBLP:conf/slsp/Bengio13}. However, these benefits are yet to be realized and evaluated in text domain.

In this work we take a representative set of \emph{unsupervised} disentanglement learning frameworks widely used in image domain (\S \ref{sec:model}) and apply them to two artificially created corpora with known underlying generative factors (\S \ref{sec:dataset}). Having known generative factors (while being ignored during the training phase) allows us to evaluate the performance of these models on imposing representation disentanglement via 6 disentanglement metrics (\S \ref{sec:sixmetrics}; \S \ref{sec:score}). Additionally, taking the highest scoring models and corresponding representations, we investigate the impact of representation disentanglement on two downstream text classification tasks (\S \ref{sec:classification}), and dimension-wise homotopy (\S \ref{sec:generation}). 

We show that existing disentanglement models, when evaluated on a wide range of metrics, are inconsistent and highly sensitive to model initialisation. However, where disentanglement is  achieved, it shows its positive impact on improving downstream task performance. Our work highlights the potential and existing challenges of disentanglement on text. We hope our proposed datasets, accessible description of disentanglement metrics and models, and experimental framework will set the path for developments of models specific to for text.
\section{Disentanglement Models and Metrics}
Let $\mathbf{x}$ denote data points and $\mathbf{z}$ denote latent variables in the latent representation space, and assume data points are generated by the combination of two random process: The first random process samples a point $\mathbf{z}^{(i)}$ from the latent space with prior distribution of $\mathbf{z}$, denoted by $p(\mathbf{z})$. The second random process generates a point $\mathbf{x}^{(i)}$ from the data space, denoted by $p(\mathbf{x}|\mathbf{z}^{(i)})$. 

We consider $\mathbf{z}$ as a disentangled representation for $\mathbf{x}$, if the changes in single latent dimensions of $\mathbf{z}$ are sensitive to changes in single generative factors of $\mathbf{x}$ while being relatively invariant to changes in other factors~\cite{Bengio2013}. Several probabilistic models are designed to reveal this process, here we look at some of the most widely used ones.

\subsection{Disentanglement Models}\label{sec:model}
A prominent approach for learning disentangled representations is through adjusting Variational Auto-Encoders (VAEs)~\citep{Kingma2014AutoEncodingVB} objective function, which decompose the representation space into independently learned coordinates. We start by introducing vanilla VAE, and then cover some of its widely used extensions that encourage disentanglement: 

\paragraph{VAE}  uses a combination of a probabilistic encoder $q_\phi(\mathbf{z}|\mathbf{x})$ and decoder $p_\theta(\mathbf{x}|\mathbf{z})$, parameterised by $\phi$ and $\theta$, to learn this statistical relationship between $\mathbf{x}$ and $\mathbf{z}$. The VAEs are trained by maximizing the lower bound of the logarithmic data distribution $\log p(\mathbf{x})$, called evidence lower bound,
\begin{equation*}\label{eq:elbo}
\resizebox{0.88\linewidth}{!}{
  \begin{minipage}{\linewidth}
\centering
$
\begin{aligned}
\mathbb{E}_{q_{\phi}(\mathbf{z} | \mathbf{x})}\big[\log p_{\theta}(\mathbf{x} | \mathbf{z})\big] - \mathbb{D}_{KL}(q_{\phi}(\mathbf{z}|\mathbf{x}),p(\mathbf{z}))
\end{aligned}
$
\end{minipage}}
\end{equation*}
The first term of is
the expectation of the logarithm of data likelihood under the posterior distribution of $z$. The second term is KL-divergence, measuring the distance between the posterior distribution $q_\phi(\mathbf{z}|\mathbf{x})$ and the prior distribution $p(\mathbf{z})$ and can be seen as a regularisation.

\paragraph{$\mathbf{\beta}$-VAE}~\citep{higgins2016} adds a hyperparameter $\beta$ to control the regularisation from the KL-term via the following objective function:
\begin{equation*}\label{eq:betavae}
\resizebox{0.88\linewidth}{!}{
  \begin{minipage}{\linewidth}
\centering
$
\begin{aligned}
\mathbb{E}_{q_{\phi}(\mathbf{z} | \mathbf{x})}\big[\log p_{\theta}(\mathbf{x} | \mathbf{z})\big] - \beta\mathbb{D}_{KL}(q_{\phi}(\mathbf{z}|\mathbf{x}),p(\mathbf{z}))
\end{aligned}
$
\end{minipage}}
\end{equation*}
Reconstructing under $\beta$-VAE (with the right value of $\beta$) framework encourages encoding data points on a set of representational axes on which nearby points along those dimensions are also close in original data space~\cite{burgess2018}. 

\paragraph{CCI-VAE}~\citep{burgess2018} extends $\beta$-VAE via constraint optimisation:
\begin{equation*}
\resizebox{0.88\linewidth}{!}{
  \begin{minipage}{\linewidth}
\centering
$
\begin{aligned}
\mathbb{E}_{q_{\phi}(\mathbf{z} | \mathbf{x})}\big[\log p_{\theta}(\mathbf{x} | \mathbf{z})\big] - \beta\left|\mathbb{D}_{KL}(q_{\phi}(\mathbf{z}|\mathbf{x}),p(\mathbf{z}))-C\right|
\end{aligned}
$
\end{minipage}}
\end{equation*}
where $C$ is a positive real value which represents the target KL-divergence term value. 
This has an information-theoretic interpretation, where the placed constraint $C$ on the KL term is seen as the amount of information transmitted from a sender (encoder) to a receiver (decoder) via the message~($\mathbf{z}$)~\cite{alemi2018fixing}, and impacts the sharpness of the posterior distribution~\cite{prokhorov-etal-2019-importance}. This constraint allows  the model to prioritize underlying factors of data according to the availability of channel capacity and their contributions to the reconstruction loss improvement.

\paragraph{MAT-VAE}~\cite{pmlr-v97-mathieu19a} introduces an additional term to $\beta$-VAE,  $\mathbb{D}_{MMD}(q_{\phi}(\mathbf{z}),p_\theta(\mathbf{z}))$,
\begin{equation*}
\resizebox{0.88\linewidth}{!}{
  \begin{minipage}{\linewidth}
\centering
$
\begin{split}
\mathbb{E}_{q_\phi(\mathbf{z}|\mathbf{x})}[\log p_\theta(\mathbf{x}|\mathbf{z})]   -
\beta \mathbb{D}_{KL}(q_{\phi}(\mathbf{z}|\mathbf{x}),p(\mathbf{z})) \\- \lambda \mathbb{D}_{MMD}(q_{\phi}(\mathbf{z}),p(\mathbf{z}))
\end{split}
$
\end{minipage}}
\end{equation*}
where $\mathbb{D}_{MMD}$ is computed using maximum mean discrepancy (\citet{JMLR:v13:gretton12a}, MMD) and $\lambda$ is the scalar weight. This term  regularises the aggregated posterior $q_{\phi}(\mathbf{z})$  with a factorised spike-and-slab prior \cite{10.2307/2290129}, which aims for disentanglement via clustering and sparsifying the representations of $\mathbf{z}$.

\subsubsection{Issue of KL-Collapse}
In text modelling, the presence of powerful auto-regressive decoders poses a common optimisation challenge for training VAEs called posterior collapse, where the learned posterior distribution $q_\phi(\mathbf{z}|\mathbf{x})$, collapses to the prior  $p(\mathbf{z})$. Posterior collapse results in
the latent variables $\mathbf{z}$ being ignored by the decoder. Several strategies have been proposed to alleviate this problem from different angles such as choice of decoders~\citep{Yang2017,bowman2015}, adding more dependency between encoder and decoder~\citep{dieng-et-al-2019-avoiding}, adjusting the training process \citep{bowman2015,he2019}, imposing direct constraints to the KL term~\citep{pelsmaeker-aziz-2020-effective,razavi2019, burgess2018,higgins2016}. In this work, both $\beta$-VAE (with $\beta<1$) and CCI-VAE are effective methods to avoid KL-collpase.

\subsection{Disentanglement Metrics}\label{sec:sixmetrics}
In this section we provide a short overview of six widely used disentanglement metrics, highlighting their key differences and commonalities, and refer the readers to the corresponding papers for exact details of computations.

\citet{eastwood2018a} define three criteria for disentangled representations: \emph{disentanglement}, which measures the degree of one dimension only encoding information about no more than one generative factor; \emph{completeness}, which measures whether a generative factor is only captured by one latent variable; \emph{informativeness}, which measures the degree by which representations capture exact values of the generative factors.\footnote{These criteria are referred to modularity, compactness and explicitness by \citet{Ridgeway2018}.} They design a series of classification tasks to predict the value of a generative factor based on the latent code, and extract the relative importance of each latent code for each task 
to calculate disentanglement and completeness scores. 
Informativeness score is measured by the accuracy of the classifier directly.  Other existing metrics reflect at least one of these three criteria, as summarised in Table~\ref{tab:metric}.

\citet{higgins2016} focus on disentanglement and propose to use the absolute difference of two groups of representations with the same value on one generative factor to predict this generative factor. For perfectly disentangled representations, latent dimensions not encoding information about this generative factor would have zero difference. Hence, even simple linear classifiers could easily identify the generative factors based on the changes of values. \citet{pmlr-v80-kim18b} consider both disentanglement and completeness by first finding the dimension which has the largest variance when fixing the value on one generative factor, and then using the found dimension to predict that generative factor. \citet{kumar2018variational} propose a series of classification tasks each of which uses a single latent variable to predict the value of a generative factor and treat the average of the difference between the top two accuracy scores for each generative factor as the final disentanglement score. 

Apart from designing classification tasks for disentanglement evaluation, another method is based on estimating the mutual information (MI) between a single dimension of the latent variable and a single generative factor. \citet{Chen2018isolating} propose to use the average of the gap (difference) between the largest normalised MI (by the information entropy of the generative factor) and the second largest normalised MI over all generative factors as the disentanglement score, whereas the modularity metric of \citet{Ridgeway2018} measures whether a single latent variable has the highest MI with only one generative factor and none 
with others.
\begin{table}[t]
\footnotesize
    \centering
    \scalebox{0.8}{
    \begin{tabular}{l c c c : c c}
        \toprule
        Metric & Dis. & Com. & Info. & Ex.1$\uparrow$& Ex.2$\uparrow$\\
        \hline
        \citet{higgins2016} & Yes & No & No&100 & 100\\
        \citet{Ridgeway2018} & Yes & No & No&100 & 100\\
        \citet{pmlr-v80-kim18b} & Yes & Yes & No&100 & 100\\
        \citet{Chen2018isolating} & No & Yes & No&81.05 & 5.73\\
        \citet{eastwood2018a} & Yes & Yes & Yes&66.47 & 63.45\\
        \citet{kumar2018variational} & No & Yes & Yes&4.68 & 3.98\\
        \bottomrule
    \end{tabular}}
    \caption{The disentanglement (Dis.), completeness (Com.), and informativeness (Info.) criteria reflected in six metrics. The Ex.1 and Ex.2 columns are corresponding metrics' scores (\%) on two ideally disentangled representations.}
    \label{tab:metric}
\end{table}

{The algorithmic details for computing the above metrics are provided in Appendix~\ref{app:alg}.}

\paragraph{Empirical Difference.} 
To highlight the empirical difference between these metrics, we use a toy set built by permuting four letters: A B C D. Each letter representing a generative factor with 20 choices of assignments (i.e, $X=\{X1,\ldots,X20\}$ where $X\in\{A,B,C,D\}$). We consider two settings where each generative factor is embedded in a single dimension (denoted by Ex.1), or two dimensions (denoted by Ex.2). In each setting we 
uniformly sample 20 values from -1 to 1 to represent 20 assignments per factor and use them to allocate the assignments into distinctive bins per each corresponding dimension. By concatenating dimensions for each generative factor, we construct two ideal disentangled representations for data points in this toy dataset, amounting to 4 and 8 dimensional representations, respectively. Using these representations (skipping the encoding step), we measured the above metrics.
Table~\ref{tab:metric} (Ex.1 and Ex.2 columns) summarises the results, illustrating that out of the 6 metrics, \citet{higgins2016, Ridgeway2018, pmlr-v80-kim18b} are the only ones that reach the potential maximum (i.e., 100), while \citet{Chen2018isolating} exhibits its sensitivity towards \emph{completeness} when we allocate two dimensions per factors.

\paragraph{Data Requirement.}
Measuring the mentioned disentanglement metrics requires a dataset satisfying the following attributes:
\begin{enumerate}[noitemsep,leftmargin=5.5mm,topsep=0pt,parsep=0pt,partopsep=0pt]
    \item A set $\mathbb{F}$ where each of its elements is a generative factor which should be disentangled through representations;
    \item For each element $f_i \in\mathbb{F}$, a value space $\mathbb{V}_i$ which is the domain of $f_i$;
    \item For each value $v_{ij}\in\mathbb{V}_i$, a sample space $\mathbb{S}_{ij}$ which contains observations who has value $v_{ij}$ on generative factor $f_i$ while everything else is arbitrary.
\end{enumerate}
We present two synthetic datasets (\S\ref{sec:dataset}) that meet these criteria and use them in our experiments (\S\ref{sec:experiments}). 

\section{Generative Synthetic Datasets}\label{sec:dataset}
The use of synthetic datasets is the common practice for evaluating disentanglement in image domain~\citep{dittadi2021on,higgins2016,pmlr-v80-kim18b}. Generative simplistic datasets in image domain define independent generative factors (e.g. shape, color) behind the data generation. However, a comparable resource is missing in text domain. We develop two synthetic generative datasets with varying degrees of  difficulty to analyse and measure disentanglement: 
The YNOC dataset (\S\ref{sec:ynoc}) which has only three structures and generative factors appearing in every sentence, and the POS dataset (\S \ref{sec:pos}) which has more structures while some generative factors are not guaranteed to appear in every sentence. The YNOC dataset offers a simpler setting for disentanglement.

\subsection{YNOC Dataset}\label{sec:ynoc}
Sentences in YNOC are generated  by 4 generative factors: Year (Y), Name (N), Occupation (O), and City (C), describing the occupation of a person. Since we often use different means to express the same message, we considered three templates to generate YNOC sentences: 
\begin{description}[noitemsep,leftmargin=-2.5mm]
\item[\ \ \ \ \ \ \ \ \textbf{Template I.}] \emph{in Y, N was a/an O in C.}
\item[\ \ \ \ \ \ \ \ \textbf{Template II.}] \emph{in Y's C, N was a/an O.}
\item[\ \ \ \ \ \ \ \ \textbf{Template III.}] \emph{N was a/an O in C in Y.}
\end{description}
The templates were then converted into real sentences using 10 years, 40 names, 20 occupations, and 30 cities. This amounted to a total of 720K sentences, split as (60\%,20\%,20\%) into training, validation, and test sets.
\begin{table}[t]
\setlength{\tabcolsep}{8pt}
    \centering
    \footnotesize
    \begin{tabular}{lc}
        \toprule
        Simple Sentence Structures & \# of Sentences\\
        \toprule
        n. v. n. end-punc. & 200\\
        n. v. adj. n. end-punc. & 1,000\\
        n. adv. v. n. end-punc. & 1,000\\
        n. adv. v. adj. n. end-punc. & 5,000\\
        n. v. prep. n. end-punc. & 1,000\\
        n. v. prep. adj. n. end-punc. & 5,000\\
        n. adv. v. prep. n. end-punc. & 5,000\\
        n. adv. v. prep. adj. n. end-punc. & 25,000\\
        adj. n. v. n. end-punc. & 1,000\\
        adj. n. v. adj. n. end-punc. & 4,000\\
        adj. n. adv. v. n. end-punc. & 5,000\\
        adj. n. adv. v. adj. n. end-punc. & 20,000\\
        adj. n. v. prep. n. end-punc. & 5,000\\
        adj. n. v. prep. adj. n. end-punc. & 20,000\\
        adj. n. adv. v. prep. n. end-punc. & 25,000\\
        adj. n. adv. v. prep. adj. n. end-punc. & 100,000\\
                \specialrule{.4em}{.3em}{.3em}
         \multicolumn{2}{l}{\textbf{n.}  [dogs cats foxes horses tigers]}\\
        \multicolumn{2}{l}{\textbf{v.}  [want need have get require]}\\
        \multicolumn{2}{l}{\textbf{adv.}  [really recently gradually frequently eventually]}\\
        \multicolumn{2}{l}{\textbf{adj.}  [happy big small beautiful fantastic]}\\
        \multicolumn{2}{l}{\textbf{prep.}  [on in for to of]}\\
        \multicolumn{2}{l}{\textbf{conj1.}  [although because when where whereas]}\\
        \multicolumn{2}{l}{\textbf{conj2.}  [and or]}\\
        \multicolumn{2}{l}{\textbf{comma} [,]}\\
        \multicolumn{2}{l}{\textbf{end-punc.} [. !]}\\
\bottomrule
    \end{tabular}
    \caption{Simple sentence structures and the vocabulary used for each POS tag in our synthetic dataset.}
    \label{tab:toy_sim}
\end{table}
\subsection{POS Dataset} \label{sec:pos}
We use part-of-speech (POS) tags to simulate the structure of sentences and define a base grammar as \emph{``n. v. n. end-punc.''}, where `n.' denotes noun, `v.' denotes verb and `end-punc.' denotes the punctuation which appears at the end of sentences. Then we define simple sentence structures as \emph{``(adj.) n. (adv.) v. (prep.) (adj.) n. end-punc.''}, where `adj.' denotes adjective, `adv.' denotes adverb, `prep.' denotes preposition, and `()' marks the arbitrary inclusion/removal of the corresponding POS tag. We populate the structures with $2^4=16$ simple structures presented in Table~\ref{tab:toy_sim}.

Next, we define complex sentence structures as  combinations of two simple sentence structures by applying one of the following three rules:

\begin{description}[noitemsep,leftmargin=-2.5mm]
\item[\ \ \ \ \ \ \ \textbf{Rule I.}] \emph{conj1. S1 comma S2 end-punc.}
\item[\ \ \ \ \ \ \ \textbf{Rule II.}] \emph{S1 conj1. S2 end-punc.}
\item[\ \ \ \ \ \ \ \textbf{Rule III.}] \emph{S1 comma conj2. S2 end-punc.}
\end{description}
\noindent where `conj1.' and `conj2.' denote two different kinds of conjunction, `comma' denotes `,' and `S1' and `S2' are two simple sentence structures without `end-punc.' We limit the number of POS tags that appear in `S1' and `S2' to 9 to control the complexity of generating sentences and obtain 279 complex structures in total. A maximum of 5 words is chosen for each POS to construct our sentences. 

The frequency of appearance for each word in a sentence is limited to one. Although this construction does not focus on sentences being ``realistic'', it simulate natural text in terms of the presence of an underlying grammar and rules over POS tags.\footnote{For structures which can produce more than 10k sentences (e.g. longer structures), we randomly choose 10k.} We deliberately ignore semantics,
since isolating semantics in terms of generative factors potentially involves analysis over multiple dimensions (combinatorial space) and quantifying grouped disentanglement requires suitable disentanglement metrics to be developed. We leave further exploration of this to our future work.

We split the dataset into training, validation and test sets with proportion 60\%, 20\%, 20\%. This proportion is used for every structure to ensure they have representative sentences in each portion of the data splits. The final size of (training, validation, test) sets are (1723680, 574560, 574560). All three sets are unbiased on word selection for each POS tag: e.g., all 5 noun POS vocabs from Table~\ref{tab:toy_sim} have equal frequency (i.e., 20\%). Exactly the same proportions are preserved for validation and test sets. 

Through the process of the generation, we can define each POS tag as one ground truth generative factor for sentences.\footnote{{While we consider POS tags as the generative factors in this paper, further sub-categorisation of POS tags based on position (e.g., first-noun and second-noun, etc) or grammatical roles (e.g., subject-noun and object-noun, etc) is a possibility for future investigation.}} Because the choices of words for different POS tags are independent, these generative factors are independent. However, for the same POS, the choices of words are dependent and POS tags are dependent on the structures as well. It is noteworthy that in contrast to the image domain where all generative factors are always present in the data, in POS  dataset this cannot be guaranteed, making it a more challenging setting.

\section{Experiments and Analysis}\label{sec:experiments}
\begin{table*}[t]
\footnotesize
    \centering
    \scalebox{0.9}{
    \begin{tabular}{l c c c c c@{\extracolsep{4pt}} c c c c}
        \toprule
        &  \multicolumn{4}{c}{YNOC} && \multicolumn{4}{c}{POS}\\
        \cline{2-5}\cline{7-10}
        Model & KL & Rec.$\downarrow$& AU$\uparrow$ & Top-3$\uparrow$& & KL & Rec.$\downarrow$& AU$\uparrow$&Top-3$\uparrow$\\
        \hline
        AE & - & $8.87_{\pm0.66}$ & $4.0_{\pm0.0}$ & 1 & &- & $4.91_{\pm1.83}$ & $8.0_{\pm0.0}$& 3\\
        Vanilla-VAE & $0.02_{\pm0.02}$ & $13.48_{\pm0.02}$ & $0.4_{\pm0.5}$& 0 & &$0.01_{\pm0.00}$ & $19.57_{\pm0.00}$ & $0.2_{\pm0.4}$& 3\\
        $\beta$-VAE ($\beta=0.2$) & $4.25_{\pm0.31}$ & $9.72_{\pm0.25}$ & $1.0_{\pm0.0}$ & 3 & & $11.19_{\pm2.88}$ & $12.03_{\pm2.04}$ & $2.8_{\pm0.7}$& 3\\
        $\beta$-VAE ($\beta=0.4$) & $3.44_{\pm0.23}$ & $10.32_{\pm0.23}$ & $1.2_{\pm0.4}$ & 1 && $7.75_{\pm0.69}$ & $13.87_{\pm0.85}$ & $2.6_{\pm0.5}$&3\\
        $\beta$-VAE ($\beta=0.8$) & $1.39_{\pm0.41}$ & $12.14_{\pm0.40}$ & $1.0_{\pm0.0}$ & 1 && $5.61_{\pm0.78}$ & $14.26_{\pm0.72}$ & $1.8_{\pm0.4}$&1\\
        CCI-VAE ($C=5$) & $5.00_{\pm0.00}$ & $9.51_{\pm0.30}$ & $1.8_{\pm1.0}$ & 1 & & $5.04_{\pm0.03}$ & $15.01_{\pm0.30}$ & $2.2_{\pm0.4}$&0\\
        CCI-VAE ($C=10$) & $10.00_{\pm0.00}$ & $9.48_{\pm0.49}$ & $3.4_{\pm0.5}$ & 2 & & $10.01_{\pm0.01}$ & $12.76_{\pm1.18}$ & $4.0_{\pm1.3}$&1\\
        MAT-VAE ($\beta=0.1,\lambda=0.1$) & $6.11_{\pm0.39}$ & $9.49_{\pm0.17}$& $1.0_{\pm0.0}$& 2 & & $22.14_{\pm2.92}$ & $8.47_{\pm2.28}$ & $3.0_{\pm0.0}$& 3\\
        MAT-VAE ($\beta=0.01,\lambda=0.1$) & $15.38_{\pm1.86}$ & $7.12_{\pm0.32}$& $3.2_{\pm0.7}$& \textbf{7} & & $45.48_{\pm1.65}$ & $3.47_{\pm0.99}$& $8.0_{\pm0.0}$ & 1\\
        \bottomrule
    \end{tabular}}
    \caption{Results are calculated on the test set. We report mean value and standard deviation across 5 runs.}
    \label{tab:loss}
\end{table*}
\begin{figure*}[t]
    \centering
    \includegraphics[width=\textwidth]{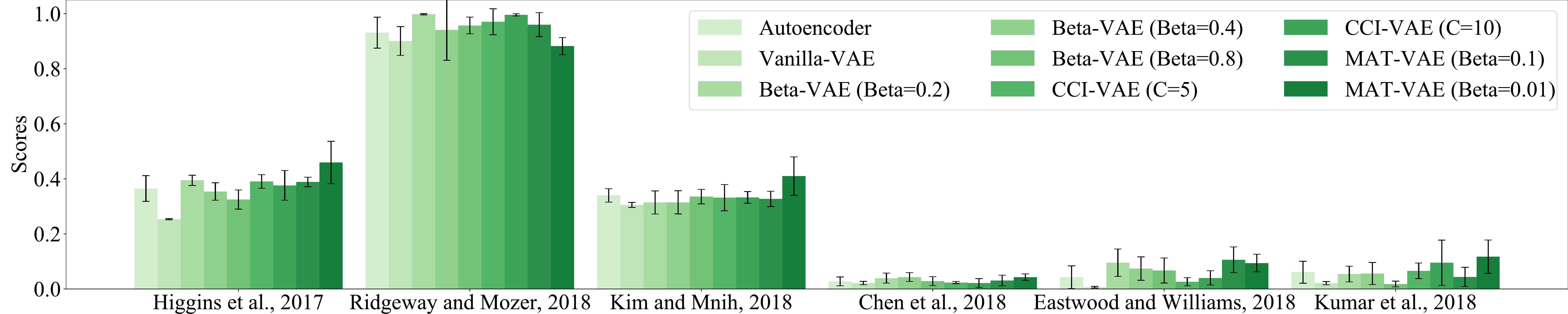}
    \includegraphics[width=\textwidth]{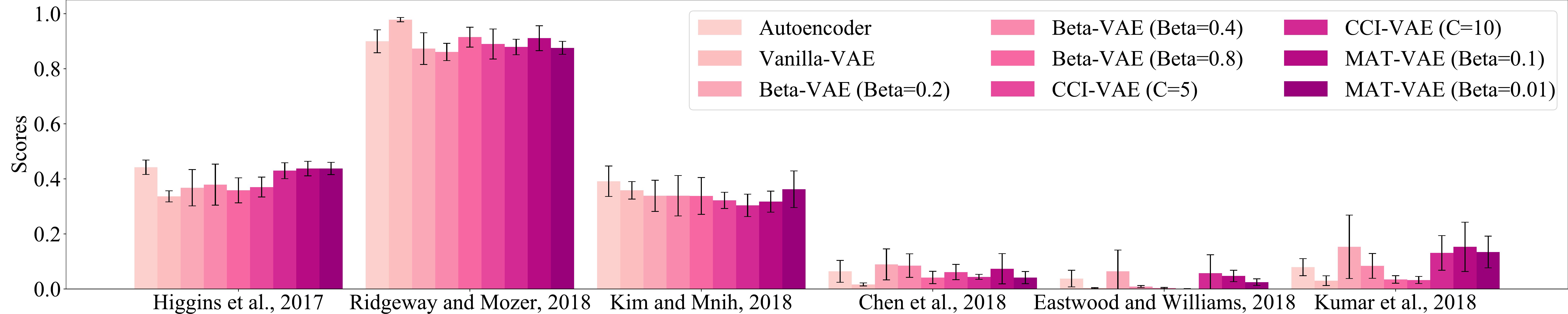}
    \caption{Disentanglement scores across six metrics on \textbf{top:} YNOC dataset and \textbf{bottom:} POS dataset. For better illustration, we multiply the scores of \citet{eastwood2018a} and \citet{kumar2018variational} by 10.}
    \label{fig:scores}
\end{figure*}
\begin{table*}[t]
\footnotesize
    \centering
    \scalebox{0.85}{
    \begin{tabular}{l c c c c c c c c c}
        \toprule
        & AE & VAE & \multicolumn{3}{c}{$\beta$-VAE} & \multicolumn{2}{c}{CCI-VAE} & \multicolumn{2}{c}{MAT-VAE}\\
        & & & $\beta=0.2$& $\beta=0.4$ & $\beta=0.8$ & $C=5$ & $C=10$& $\beta=0.1,\lambda=0.1$ & $\beta=0.01,\lambda=0.1$\\
                \cmidrule(lr){2-2}\cmidrule(lr){3-3}\cmidrule(lr){4-6}\cmidrule(lr){7-8}\cmidrule(lr){9-10}
        YNOC & $0.22_{\pm0.03}$ & $0.03_{\pm0.02}$ & $0.30_{\pm0.03}$ & $0.30_{\pm0.02}$ & $0.30_{\pm0.05}$ & $0.32_{\pm0.04}$ & $0.30_{\pm0.01}$ & $0.36_{\pm0.03}$ & $0.43_{\pm0.09}$\\
        POS & $0.30_{\pm0.05}$ & $0.21_{\pm0.03}$ & $0.25_{\pm0.00}$ & $0.27_{\pm0.01}$ & $0.29_{\pm0.04}$ & $0.29_{\pm0.05}$ & $0.28_{\pm0.01}$ & $0.29_{\pm0.00}$ & $0.28_{\pm0.01}$\\
        \bottomrule
    \end{tabular}}
    \caption{Hoyer scores are calculated on the test set. We report mean value and standard deviation across 5 runs.}
    \label{tab:hoyer}
\end{table*}
In this section, we examine the introduced disentanglement models on text. We measure the disentanglement scores of each model on our two synthetic datasets and quantify how well-correlated these metrics are with reconstruction loss, active units, and KL~(\S \ref{sec:score}). We then look at various strategies for coupling the latent code during decoding and highlight their impacts on training and disentanglement behaviors (\S \ref{sec:code}). We continue our analysis by showing how the representation learned by the highest scoring model (on  disentanglement metrics) performs compared to vanilla VAE in two text classification tasks (\S \ref{sec:classification}), and finish our analysis by looking at these models' generative behaviors (\S \ref{sec:generation}).
\paragraph{Training Configuration.} We adopt the VAE architecture from \cite{bowman2015}, using a LSTM encoder-decoder. Unless stated otherwise, (word embedding, LSTM, representation embedding) dimensionalities for YNOC and POS datasets are (4D, 32D, 4D) and (4D, 64D, 8D), respectively, and we use the latent code to initialize the hidden state of the LSTM decoder. We use greedy decoding. All models are trained from multiple random starts using Adam~\citep{kingma2014} with learning rate 0.001 for 10 epochs. We set batch size to 256 and 512 for YNOC and POS, respectively.

\subsection{Disentanglement Metrics}\label{sec:score}
Taking the models (\S\ref{sec:model}) and also an Autoencoder (AE) as a baseline we use the YNOC and POS datasets to report average KL-divergence (KL), reconstruction loss (Rec.), and number of active units (AU)\footnote{$i$ is active if  ${\rm Covariance}_{\mathbf{x}}(\mathbb{E}_{i\sim q\left(i|\mathbf{x}\right)}\left[i\right])>0.01$.} in Table~\ref{tab:loss}, and illustrate disentanglement metrics' scores in Figure~\ref{fig:scores}. 

As demonstrated in Table~\ref{tab:loss}, different models pose various behaviors, noteworthy of those are: (1) the positive correlation of $C$ with AU which intuitively means the increase of channel capacity demands more dimensions of the representation to carry information which then translates into having a better reconstruction of data, (2) the negative correlation between the increase of $\beta$ and decrease of reconstruction loss, (3) the best Rec. and AU are achieved by AE and MAT-VAE whereas the worst one is achieved by the (collapsed) vanilla-VAE, (4) the MAT-VAE ($\beta=0.01,\lambda=0.1$) model which induces more sparse representations \footnote{Sparsity is measured using Hoyer \cite{Hoyer_metric}. In this paper we report this as the average Hoyer over data points' posterior means. Hoyer for data point $x_i$ with posterior mean $\mu_i$ is calculated as $\frac{\sqrt{d} - ||\bar{\mu}_i||_1/||\bar{\mu}_i||_2 }{\sqrt{d} -1},$ where $d$ is the dimensionality of the representations and $\bar{\mu}_i = \mu_i / \sigma(\mu)$, where $\mu = \{ \mu_1,...,\mu_n \}$, and $\sigma(.)$ is the standard deviation.} performs the best on both datasets, indicating the positive impact of representation sparsity as an inductive bias.

As illustrated in Figure~\ref{fig:scores}, the difference between means of each disentanglement score on various models is relatively small, and due to large standard deviation on metrics, it is difficult to single out a superior model. This verifies findings of~\citet{pmlr-v97-locatello19a} on image domain. In Table~\ref{tab:loss} (Top-3 column) we report the number of appearances of a model among the top 3 highest scoring models on at least one disentanglement metric. 
%
The ranking suggests that $\beta$-VAE with smaller $\beta$ values reach better disentangled representations, and MAT-VAE performing superior on YNOC and poorly on POS, highlighting its more challenging nature. For MAT-VAE we also observe an interesting correlation between sparsity and disentanglement: for instance on YNOC, MAT-VAE ($\beta=0.01,\lambda=0.1$) achieves the highest Hoyer (See Table~\ref{tab:hoyer}) and occurs 7 times among Top-3 (see Table~\ref{tab:loss}). Interestingly, the success of MAT-VAE does not translate to POS dataset, where it underperforms AE. These two observations suggest that sparsity could be a facilitator for disentanglement, but achieving a stable level of sparsity remains as a challenge. The more recent development in the direction of sparsity, HSVAE~\cite{prokhorov2020hierarchical}, addresses the  stability issue of MAT-VAE but we leave its exploration to future work.
\begin{figure}
    \centering
    \includegraphics[trim=0 0 9cm 0, clip,scale=0.18]{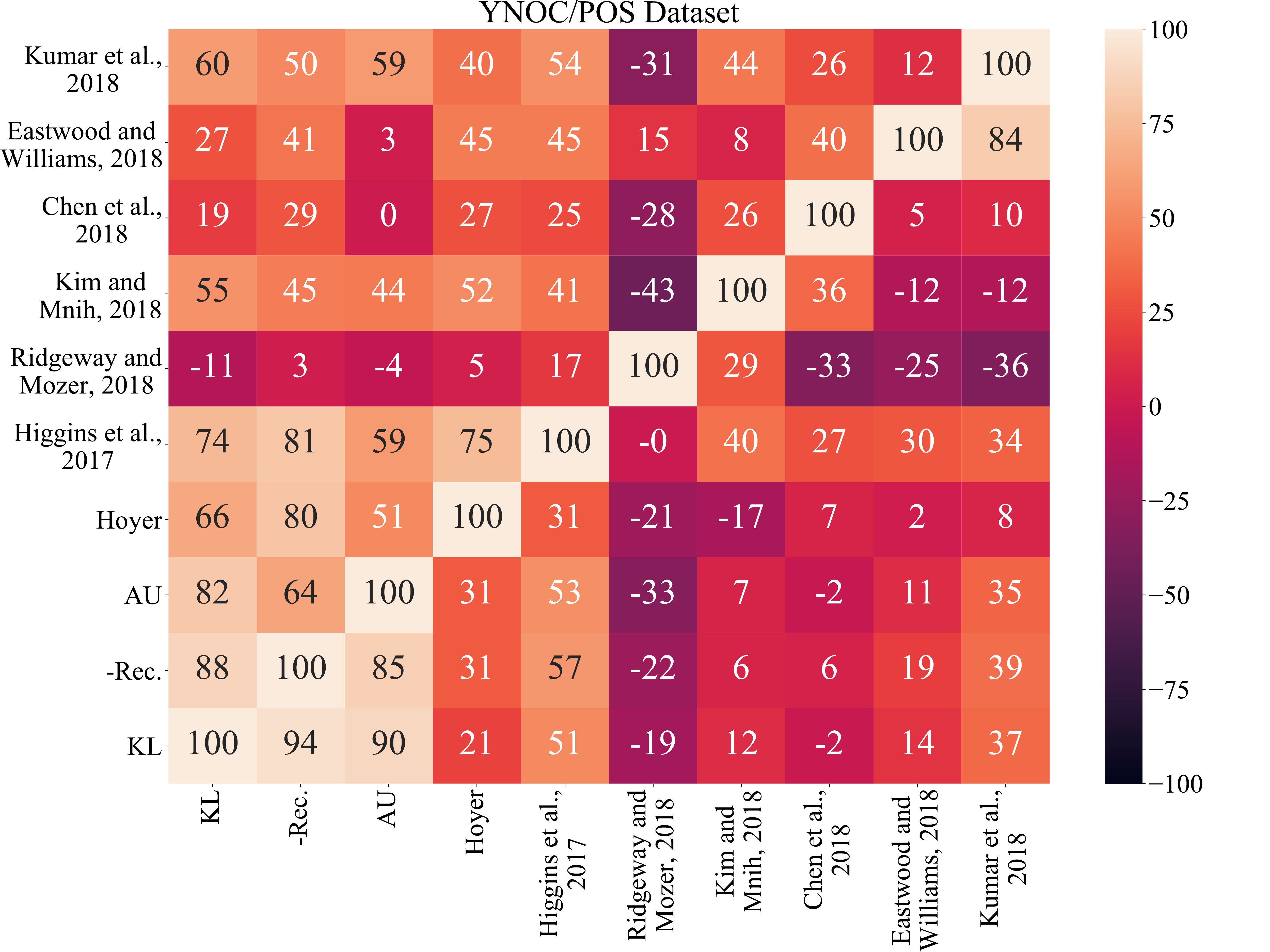}
    \caption{Correlation coefficients between six disentanglement metrics, Hoyer, AU, Rec, and KL on \textbf{Upper Triangle:} YNOC dataset and \textbf{Lower Triangle:} POS dataset.}
    \label{fig:corr}
\end{figure}

\begin{figure}
\centering
\begin{subfigure}[b]{0.48\textwidth}
   \includegraphics[width=1\linewidth]{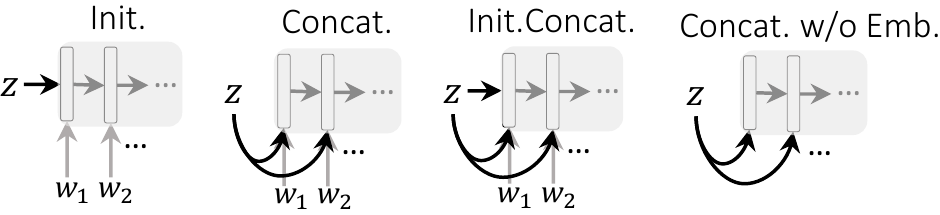}
   \caption{Different coupling strategies for the latent code and decoder (\S\ref{sec:code}). Gray box denotes decoder.}
   \label{fig:couplingfig} 
\end{subfigure}
\begin{subfigure}[b]{0.48\textwidth}
   \includegraphics[width=1\linewidth]{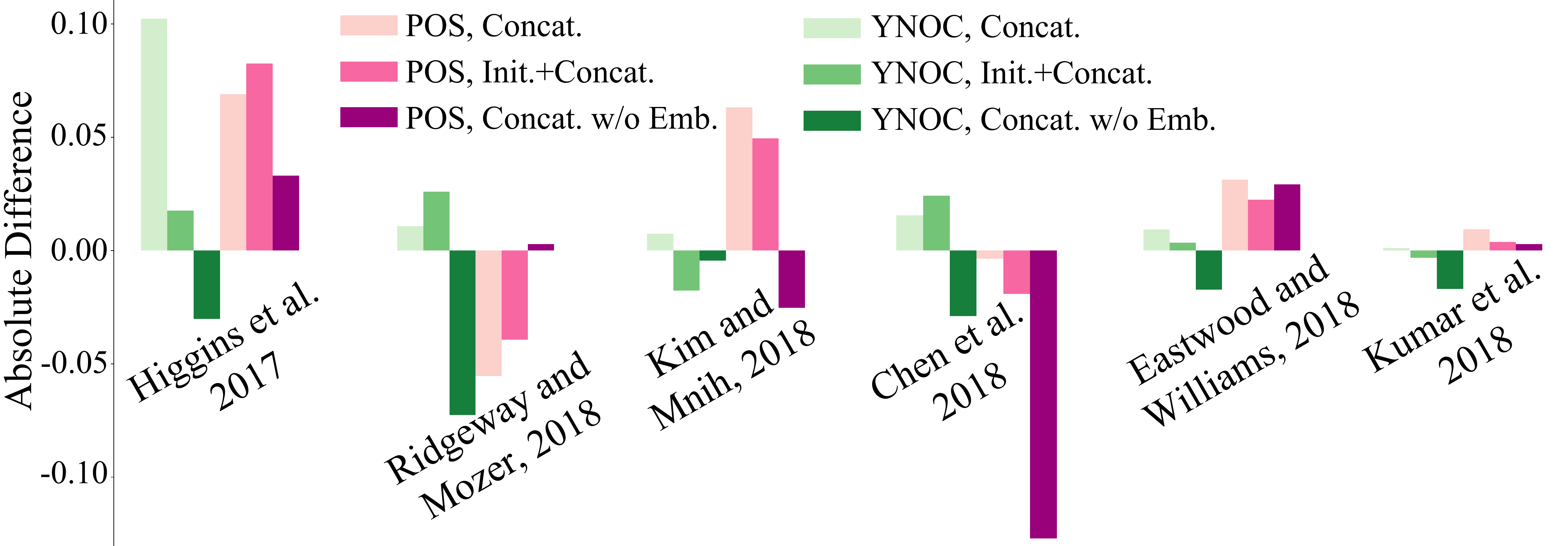}
   \caption{Absolute differences between disentanglement metrics' scores of Init. coupling and others (\S\ref{sec:code}).}
   \label{fig:diff}
\end{subfigure}
\caption{Different coupling strategies for the latent code and decoder and their impacts on disentanglement on POS and YNOC.}
\end{figure}
\begin{table}
\footnotesize
    \centering
    \scalebox{0.72}{
    \begin{tabular}{ll c c c c}
        \toprule
        &&  \multicolumn{4}{c}{Coupling Methods}\\
         &&Init.&Concat.&Init.Concat.&Concat. w/o Emb.\\
         \cline{3-6}
        \multirow{3}{*}{\rotatebox[origin=c]{90}{{YNOC}}} & KL  
	    &$1.51{\pm0.01}$   & $1.52{\pm0.01}$  &$1.52{\pm0.01}$   &$1.62{\pm0.04}$ \\
	    &Rec.$\downarrow$& $12.04{\pm0.04}$ &$12.06{\pm0.03}$   &$12.01{\pm0.02}$   &$12.29{\pm0.16}$ \\
	    &AU$\uparrow$&$1.2{\pm0.4}$   &$2.0{\pm0.0}$  &  $1.0{\pm0.0}$ &$1.2{\pm0.4}$ \\\cmidrule(lr){3-6}
	            \multirow{3}{*}{\rotatebox[origin=c]{90}{{POS}}} & KL  
	    &$5.54{\pm0.02}$  &$5.53{\pm0.02}$  & $5.51{\pm0.00}$  &$5.69{\pm0.03}$ \\
	    &Rec.$\downarrow$&$14.54{\pm0.33}$   & $15.89{\pm0.26}$ & $15.98{\pm0.05}$  &$16.48{\pm0.09}$ \\
	    &AU$\uparrow$& $2.2{\pm0.4}$ &$4.0{\pm0.0}$  & $3.2{\pm0.4}$ &$3.6{\pm0.5}$\\
        \bottomrule
    \end{tabular}}
    \caption{Test set KL, Reconstruction loss, Active Units using 4 coupling methods (\S \ref{sec:code}).}
    \label{tab:mode}
\end{table}

To further analyse the inconsistency between different metrics we calculate the Pearson product-moment correlation coefficient between them and KL, -Rec, AU, Hoyer on POS and YNOC datasets. See the heatmap in Figure~\ref{fig:corr}. 
While text-specific metrics are yet to be developed, our experiment suggests 
\citet{higgins2016} is a good candidate to try first for text domain as it seems to be the one with strong correlation with Hoyer, AU, -Rec, and KL and has the highest level of agreement (overall) with other metrics.

\subsection{Coupling Latent Code and Decoder}\label{sec:code}

In VAEs, we typically feed the decoder with the latent code as well as word embeddings during training. 
The method to couple the latent code with decoder could have some effects on disentanglement for text. To highlight this, we train with 4 different coupling strategies: \emph{Init},
\emph{Concat}, 
\emph{Init Concat}, 
\emph{Concat w/o Emb}.
See Figure~\ref{fig:couplingfig} for an accessible visualisation. {To analyse the impact of coupling, we opt for CCI-VAE which allows the comparisons to be made for the same value of KL.}

We first use \emph{Concat w/o Emb} to find an optimal KL in vanilla VAEs, which is then used as the $C$ to train CCI-VAEs using the other coupling metrics on YNOC and POS datasets. For YNOC, $C=1.5$, and for POS, $C=5.5$. This is to keep KL-divergence and reconstruction loss at the same level for fair comparison across different strategies. We report results in Table~\ref{tab:mode}. Among the investigated coupling methods, the key distinguishing factor for disentanglement is their impacts on AU which is the highest for \emph{Concat}. 

Next, using \emph{Init} as the baseline, we measure the absolute difference between disentanglement scores of different coupling methods in Figure~\ref{fig:diff}. In general, using concatenation can bring a large improvement in disentanglement. Using both initialization and concatenation do not lead to a better result.
Despite our expectation, not feeding word embeddings into decoder during training does not encourage disentanglement due to the added reliance on the latent code. 

A confounding factor which could pollute this analysis is the role of strong auto-regressive decoding of VAEs and the type of information captured by the decoder in such scenario. While a preliminary analysis has been provided recently~\cite{bosc-vincent-2020-sequence}, this has been vastly under-explored and requires more explicit attempts. We leave deeper investigation of this to future work.
\begin{figure}[t]
    \centering
    \includegraphics[scale=0.5]{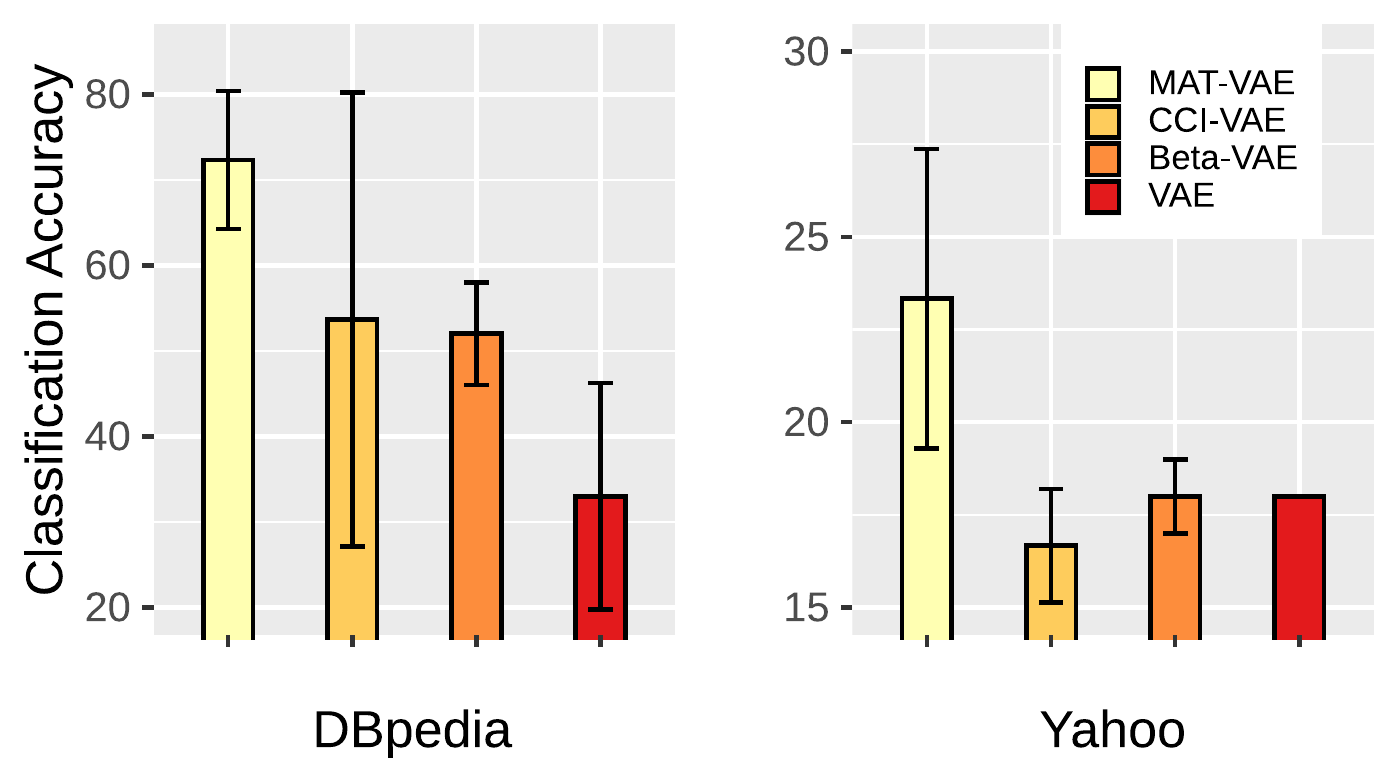}
    \caption{Classification accuracy on DBpedia and Yahoo Question using different VAE models. Results are reported as mean and std across 3 randomly initialised runs.}
    \label{fig:clf}
\end{figure}
\subsection{Disentanglement and Classification}\label{sec:classification}
To examine the performance of these models on real-world downstream task setting, we consider the classification task. For our classification datasets, we use DBpedia (14 classes) and Yahoo Question (10 classes)~\citep{10.5555/2969239.2969312}.
Each \textbf{class} of these two datasets has (10k, 1k, 1k) randomly chosen sentences in (train, dev, test) sets. We train Vanilla-VAE, $\beta$-VAE ($\beta=0.2$), CCI-VAE ($C=10$), and MAT-VAE ($\beta=0.01,\lambda=0.1$) from Table~\ref{tab:loss} on DBpedia and Yahoo (without the labels), then freeze the trained encoders and place a classifier on top to use the mean vector representations from the encoder as a feature to train a classifier.  

We set the dimensionality of word embedding, LSTM, and the latent space to 128, 512, 32, respectively. The VAE models are trained using a batch size of 64, for 6 epochs with Adam (learning rate 0.001).  For the classifier, we use a single linear layer with  1024 neurons, followed by a Softmax and train it for 15 epochs, using Adam (learning rate 0.001) and batch size 512. We illustrate the mean and standard deviation across 3 runs of models in Figure~\ref{fig:clf}. 

We observe that the ranking of classification accuracy among the models on DBpedia is consistent with their Top-3 performance in Table~\ref{tab:loss}, with MAT-VAE outperforming the other three variants. We see roughly the same trend for Yahoo, with MAT-VAE being the dominating model. This indicates that disentangled representations are likely to be easier to discriminate, although the role of sparsely learned representations could contribute to MAT-VAE's success as well~\cite{prokhorov2020hierarchical}.

\subsection{Disentanglement and Generation}\label{sec:generation}
\begin{table}
    \centering
    \scalebox{0.82}{
    \begin{tabular}{clccc}
    \toprule
         \textsc{Start}&$z_1$&$[\mathbf{z_{1,1},z_{1,2},z_{1,3}}]$& &\\ $i=1$&$z'_{1,1}$ &\solidbox{$z_{1,1}$}$,$\ {$z_{1,2},z_{1,3}$}&$\rightarrow$ &\solidbox{$z_{2,1}$}$,$\ {$z_{1,2},z_{1,3}$}\\ 
         $i=2$&$z'_{1,2}$&\dashedbox{$z_{2,1}$}$,$\solidbox{$z_{1,2}$}$,${$z_{1,3}$}&$\rightarrow$ &\dashedbox{$z_{2,1}$}$,$\solidbox{$z_{2,2}$}$,z_{1,3}$\\
         $i=3$&$z'_{1,3}$&\dashedbox{$z_{2,1},z_{2,2}$}$,$\solidbox{$z_{1,3}$}&$\rightarrow$ &\dashedbox{$z_{2,1},z_{2,2}$}$,$\solidbox{$z_{2,3}$}\\
         \textsc{End} &$z_2$&& &$[\mathbf{z_{2,1},z_{2,2},z_{2,3}}]$\\
         \bottomrule
    \end{tabular}}
    \caption{An example of a 3D latent code transformation in the dimension-wise homotopy. In row $i$, $\rightarrow$ denotes the start and end points of interpolation, solid box denotes the two dimensions being interpolated, and dashed box denotes the updated dimensions from $i-1$.}
    \label{tab:dimhomotopy}
\end{table}
To observe the effect of disentanglement in homotopy~\cite{bowman2015}, we 
use the exactly same toy dataset introduced in \S \ref{sec:model} and assess the homotopy behaviour of the highest scoring VAE vs. an ideal representation. To conduct homotopy, we interpolate between two sampled sequences' representations and pass the intermediate representations to decoder to generate the output. We use 4D word embedding, 16D LSTM, 4D latent space. We report the results for the VAEs scoring the highest on disentanglement (w.r.t. \citet{higgins2016} denoted as VAE-Higg) and completeness (w.r.t. \citet{Chen2018isolating} denoted as VAE-Chen). The VAE-Higg and VAE-Chen are $\beta$-VAE with $\beta=0.4$ and MAT-VAE with $\beta=0.01,\lambda=0.1$, respectively.

Additionally, to highlight the role of generative factor in generation, we conduct a dimension-wise homotopy, transitioning from the first to the last sentence by interpolating between the dimensions one-by-one. This is implemented as follows: (i) using prior distribution\footnote{Instead of prior, we sample two sentences from test set and use their representations. This is to avoid the situation where samples are not in the well-estimated region of the posterior.} we sample two latent codes denoted by $\textbf{z}_1=(z_{1,1},z_{1,2},\dots,z_{1,n})$, $\textbf{z}_2=(z_{2,1},z_{2,2},\dots,z_{2,n})$; (ii) for $i$-th dimension, using $\textbf{z}_{1,i}^{'}=(z_{2,1},\dots,z_{2,i-1},z_{1,i},\dots,z_{1,n})$ as the start, we interpolate along the $i$-th dimension towards $\textbf{z}_{2,i}^{'}=(z_{2,1},\dots,z_{2,i},z_{1,i+1},\dots,z_{1,n})$. Table~\ref{tab:dimhomotopy} illustrates this for a 3D latent code example.

\paragraph{Results:} Table~\ref{tab:ideal_homo} reports the outputs for standard homotopy (top block) and dimension-wise homotopy. The results for standard homotopy demonstrate that the presence of ideally disentangled representation translates into disentangled generation in general. However, both VAE-Higg and VAE-Chen seem to mainly be producing variations of the letter in the first position (letter \textbf{A}) during the interpolation. The same observation holds in the dimension-wise experiments. VAE-Chen also produces variations of the letter in the second position (letter \textbf{B}) along with the variation of letter \textbf{A}, which suggests the lesser importance of completeness for disentangled representations.

\begin{table}[t]
\footnotesize
    \centering
    \scalebox{0.76}{
    \begin{tabular}{l p{0.15\textwidth} p{0.17\textwidth} p{0.17\textwidth}}
        \toprule
        & Ideal & VAE-Higg & VAE-Chen\\
        \hline
        $\mathbf{z_1}$& \textbf{A9 B17 C13 D3} & \textbf{A12 B14 C14 D12} & \textbf{A9 B4 C10 D15}\\
        \textbf{
        \parbox[t]{4mm}{\multirow{4}{*}{\rotatebox[origin=c]{90}{Homotopy}}}} & A20 B17 C1 D3 & A12 B14 C14 D12 & A7 B4 C10 D15\\ 
        & A4 B17 C12 D6 & A8 B14 C14 D12 & A14 B4 C10 D15\\
        & A3 B1 C6 D6 & A20 B14 C14 D12 & A20 B19 C10 D15\\
        & A13 B1 C6 D20 & A15 B14 C14 D12 & A8 B19 C10 D15\\
        $\mathbf{z_2}$& \textbf{A15 B2 C8 D10} & \textbf{A4 B14 C14 D12} & \textbf{A12 B19 C10 D15}\\
        \hline
        $\textbf{z}_1$ & \textbf{A9 B17 C13 D3} & \textbf{A12 B14 C14 D12} & \textbf{A9 B4 C10 D15}\\
        \cdashlinelr{1-4}
        \textbf{
        \parbox[t]{4mm}{\multirow{3}{*}{\rotatebox[origin=c]{90}{Dim 1}}}}
        & A20 B17 C13 D3 & A12 B14 C14 D12 & A7 B4 C10 D15\\
        & A4 B17 C13 D3 & A8 B14 C14 D12 & A4 B19 C10 D15\\
        & A3 B17 C13 D3 & A20 B14 C14 D12 & A8 B19 C10 D15\\
        & A13 B17 C13 D3 & A18 B14 C14 D12 & A12 B19 C10 D15\\
        \cdashlinelr{1-4}
        $\textbf{z}_{1,2}^{'}$ & \textbf{A15} B17 C13 D3 & \textbf{A4} B14 C14 D12 & \textbf{A12} B19 C10 D15\\
        \cdashlinelr{1-4}
        \textbf{
        \parbox[t]{4mm}{\multirow{3}{*}{\rotatebox[origin=c]{90}{Dim 2}}}}
        & A15 B17 C13 D3 & A4 B14 C14 D12 & A12 B19 C10 D15\\
        & A15 B17 C13 D3 & A4 B14 C14 D12 & A12 B19 C10 D15\\
        & A15 B17 C13 D3 & A4 B14 C14 D12 & A12 B19 C10 D15\\
        & A15 B1 C13 D3 & A4 B14 C14 D12 & A12 B19 C10 D15\\
        \cdashlinelr{1-4}
        $\textbf{z}_{1,3}^{'}$ & A15 \textbf{B2} C13 D3 & A4 \textbf{B14} C14 D12 & A12 \textbf{B19} C10 D15\\
        \cdashlinelr{1-4}
        \textbf{
        \parbox[t]{4mm}{\multirow{3}{*}{\rotatebox[origin=c]{90}{Dim 3}}}}
        & A15 B2 C1 D3 & A4 B14 C14 D12 & A12 B19 C10 D15\\
        & A15 B2 C12 D3 & A4 B14 C14 D12 & A12 B19 C10 D15\\
        & A15 B2 C6 D3 & A4 B14 C14 D12 &A12 B19 C10 D15\\
        & A15 B2 C6 D3 & A4 B14 C14 D12 & A12 B19 C10 D15\\
        \cdashlinelr{1-4}
        $\textbf{z}_{1,4}^{'}$ & A15 B2 \textbf{C8} D3 & A4 B14 \textbf{C14} D12  & A12 B19 \textbf{C10} D15\\
        \cdashlinelr{1-4}
        \textbf{
        \parbox[t]{4mm}{\multirow{3}{*}{\rotatebox[origin=c]{90}{Dim 4}}}}
        & A15 B2 C8 D3 & A4 B14 C14 D12 & A12 B19 C10 D15\\
        & A15 B2 C8 D6 & A4 B14 C14 D12 & A12 B19 C10 D15\\
        & A15 B2 C8 D6 & A4 B14 C14 D12 & A12 B19 C10 D15\\
        & A15 B2 C8 D20 & A4 B14 C14 D12 & A12 B19 C10 D15\\
        \cdashlinelr{1-4}
        $\textbf{z}_{2}$ & \textbf{A15 B2 C8 D10} & \textbf{A4 B14 C14 D12} & \textbf{A12 B19 C10 D15}\\
        \bottomrule
    \end{tabular}}
    \caption{The homotopy experiments, comparing an ideal generator and the best disentangled VAEs according to \citet{higgins2016} (VAE-Higg) and \citet{Chen2018isolating} (VAE-Chen).}
    \label{tab:ideal_homo}
\end{table}
This indicates that despite the relative superior performance of certain models on  the metrics and classification tasks, the amount of disentanglement present in the representation is not sufficient enough to be reflected by the generative behavior of these models. As a future work, we would look into  the role of auto-regressive decoding and teacher-forcing as confounding factors that can potentially affect the disentanglement process.




\section{Conclusion and Future Directions}
We evaluated a set of recent \emph{unsupervised} disentanglement learning frameworks widely used in image domain on two artificially created corpora with known underlying generative factors. Our experiments highlight the existing gaps in text domain, the daunting tasks state-of-the-art models from image domain face on text, and the confounding elements that pose further challenges towards representation disentanglement in text domain. Motivated by our findings, in future, we will explore the role of inductive biases such as representation sparsity in achieving representation disentanglement. Additionally, we will look into alternative forms of decoding and training which may compromise reconstruction quality but increase the reliance of decoding on the representation, hence allowing for a more controlled analysis and evaluation.

Our synthetic datasets and experimental framework provide a set of quantitative and qualitative measures to facilitate and  future research in developing new models, datasets, and evaluation metrics specific for text.


\bibliographystyle{acl_natbib}
\bibliography{references.bib}
\appendix
\section{Disentanglement Metrics Algorithms}\label{app:alg}
To evaluate representations learned by a model on a dataset having the attributes of \textbf{Data Requirement}, we further require a series of representation space $\mathbb{R}_{ij}$, who has a bijection mapping with $\mathbb{S}_{ij}$. Hence, when sampling representations which have the same value on one generative factor, we only need to sample in one $\mathbb{R}_{ij}$.

Under these notations, we write the pseudo code of metrics in Algorithm \ref{alg:higgins}-\ref{alg:eastwood}. For Algorithm \ref{alg:ridgeway} and \ref{alg:eastwood}, although we only use one criterion in the main paper, we still provide the details for other criteria. We set $N=1000$ and $L=64$ for Algorithm \ref{alg:higgins} and \ref{alg:kim}, and $N=10000$ for Algorithm \ref{alg:kumar}, \ref{alg:chen}, \ref{alg:ridgeway}, and \ref{alg:eastwood}.

\begin{algorithm}
    \caption{Metric of \citet{higgins2016}}
    \label{alg:higgins}
    \begin{algorithmic}[1]
        \State{$\mathbb{D}=\varnothing$}
        \For{$f_i \in\mathbb{F}$}
            \For{$n=1,2,\ldots,N$}
                \State{Sample $s_n$ from $\bigcup_{j}\mathbb{S}_{ij}$}
                \State{Find the value $v_{ij}$ on $f_i$ for $s_n$}
                \State{Sample $(\mathbf{z}_1^{(1)},\ldots,\mathbf{z}_{L}^{(1)})$ from $\mathbb{R}_{ij}$}
                \State{Sample $(\mathbf{z}_1^{(2)},\ldots,\mathbf{z}_{L}^{(2)})$ from $\mathbb{R}_{ij}$}
                \State{$\mathbf{z}_{n}=\frac{1}{L}\sum_{l=1}^{L}|\mathbf{z}_l^{(1)}-\mathbf{z}_l^{(2)}|$}
                \State{$\mathbb{D}=\{(\mathbf{z}_{n},f_i)\}\bigcup\mathbb{D}$}
            \EndFor
        \EndFor
        \State{Split $\mathbb{D}$ into training set $\mathbb{TR}$ and test set $\mathbb{TE}$ with proportion (80\%, 20\%)}
        \State{Train 10 MLPs with only input and output layer on $\mathbb{TR}$}
        \State{Calculate the accuracy on $\mathbb{TE}$ for 10 models}
        \State{Calculate the mean and variance of accuracy}
    \end{algorithmic}
\end{algorithm}
\begin{algorithm}
    \caption{Metric of \citet{pmlr-v80-kim18b}}
    \label{alg:kim}
    \begin{algorithmic}[1]
        \State{$\mathbb{D}=\varnothing$}
        \For{$d=1,2,\ldots,dim_z$}
            \State{Calculate the standard deviation $\sigma_d$ of dimension $d$}
        \EndFor
        \For{$f_i \in\mathbb{F}$}
            \For{$n=1,2,\ldots,N$}
                \State{Sample $s_n$ from $\bigcup_{j}\mathbb{S}_{ij}$}
                \State{Find the value $v_{ij}$ on $f_i$ for $s_n$}
                \State{Sample $(\mathbf{z}_1,\ldots,\mathbf{z}_{L})$ from $\mathbb{R}_{ij}$}
                \State{$d^{*}_n=\mathop{\arg\max}_d var(\frac{z_{1,d}}{\sigma_d},\ldots,\frac{z_{L,d}}{\sigma_d})$}
                \State{$\mathbb{D}=\{(d^{*}_n,f_i)\}\bigcup\mathbb{D}$}
            \EndFor
        \EndFor
        \State{Split $\mathbb{D}$ into training set $\mathbb{TR}$ and test set $\mathbb{TE}$ with proportion (80\%, 20\%)}
        \State{Train 10 majority vote classifiers on $\mathbb{TR}$}
        \State{Calculate the accuracy on $\mathbb{TE}$ for 10 models}
        \State{Calculate the mean and variance of accuracy}
    \end{algorithmic}
\end{algorithm}
\begin{algorithm}
    \caption{Metric of \citet{kumar2018variational}}
    \label{alg:kumar}
    \begin{algorithmic}[1]
        \For{$f_i \in\mathbb{F}$}
            \For{$v_{ij}\in\mathbb{V}_i$}
                \State{$p(v_{ij})=\frac{{Count}(\mathbb{S}_{ij})}{\sum_{j}{Count}(\mathbb{S}_{ij})}$}
                \State{Sample $N_j=N\times p(v_{ij})$ representations $\mathbf{z}^{j}$ from $\mathbb{R}_{ij}$}
            \EndFor
            \For{$d=1,2,\ldots,dim_z$}
                \State{$\mathbb{D}_{d}=\varnothing$}
                \For{$v_{ij}\in\mathbb{V}_i$}
                    \For{$n=1,2,\ldots,N_j$}
                        \State{$\mathbb{D}_{id}=\{(z_{n,d}^{j},v_{ij})\}\bigcup\mathbb{D}_{id}$}
                    \EndFor
                \EndFor
                \State{Split $\mathbb{D}_{d}$ into training set $\mathbb{TR}_{d}$ and test set $\mathbb{TE}_{d}$ with proportion (80\%, 20\%)}
                \State{Train a linear SVM classifier on $\mathbb{TR}_{d}$}
                \State{Record the accuracy ${acc}_{d}$ on $\mathbb{TE}_{id}$}
            \EndFor
            \State{$d^{*}=\mathop{\arg\max}_d {acc}_d$}
            \State{${SAP}_i={acc}_{d^{*}}-\max_{d\neq d^{*}}{acc}_d$}
        \EndFor
        \State{${score}={avg}({SAP}_i)$}
    \end{algorithmic}
\end{algorithm}
\begin{algorithm}
    \caption{Metric of \citet{Chen2018isolating}}
    \label{alg:chen}
    \begin{algorithmic}[1]
        \For{$d=1,2,\ldots,dim_z$}
            \State{Divide values on dimension $d$ into 20 uniform bins $\mathbb{B}_d$}
            \For{$n=1,2,\ldots,20$}
                \State{ $p(z_d\in\mathbb{B}_d^{n})=\frac{{Count}(\{z_d\in\mathbb{B}_d^{n}\})}{\sum_{n=1}^{20}{Count}(\{z_d\in\mathbb{B}_d^{n}\})}$}
            \EndFor
            \State{$H(z_d)=-\sum_{n=1}^{20} p(z_d\in\mathbb{B}_d^{n})\log p(z_d\in\mathbb{B}_d^{n})$}
        \EndFor
        \For{$f_i \in\mathbb{F}$}
            \For{$v_{ij}\in\mathbb{V}_i$}
                \State{$p(v_{ij})=\frac{{Count}(\mathbb{S}_{ij})}{\sum_{j}{Count}(\mathbb{S}_{ij})}$}
                \State{Sample $N_j=N\times p(v_{ij})$ representations $\mathbf{r}^{j}$ from $\mathbb{R}_{ij}$}
            \EndFor
            \State{$H(f_i)=-\sum_{j} p(v_{ij})\log p(v_{ij})$}
            \For{$d=1,2,\ldots,dim_z$}
                \For{$v_{ij}\in\mathbb{V}_i$}
                    \For{$n=1,2,\ldots,20$}
                        \State{ $p(z_d\in\mathbb{B}_d^{n}|v_{ij})=\frac{{Count}(\{r^{j}_d\in\mathbb{B}_d^{n}\})}{\sum_{n=1}^{20}{Count}(\{r^{j}_d\in\mathbb{B}_d^{n}\})}$}
                    \EndFor
                \EndFor
                \State{$H(z_d|f_i)=-\sum_{j}p(v_{ij})\sum_{n=1}^{20} p(z_d\in\mathbb{B}_d^{n}|v_{ij})\log p(z_d\in\mathbb{B}_d^{n}|v_{ij})$}
                \State{$I(z_d,f_i)=H(z_d)-H(z_d|f_i)$}
            \EndFor
            \State{$d^{*}=\mathop{\arg\max}_d\frac{I(z_d,f_i)}{H(f_i)}$}
            \State{${MIG}_i=\frac{I(z_{d^{*}},f_i)}{H(f_i)}-\max_{d\neq d^{*}}\frac{I(z_{d},f_i)}{H(f_i)}$}
        \EndFor
        \State{${score}={avg}({MIG}_i)$}
    \end{algorithmic}
\end{algorithm}
\begin{algorithm}
    \caption{Metric of \citet{Ridgeway2018}}
    \label{alg:ridgeway}
    \textbf{Modularity:}
    \begin{algorithmic}[1]
        \State{Same steps as Algorithm \ref{alg:chen} without step 17, 18 and 19}
        \For{$d=1,2,\ldots,dim_z$}
            \State{$i^{*}=\mathop{\arg\max}_iI(z_d,f_i)$}
            \State{$\theta_d=I(z_d,f_{i^{*}})$}
            \For{$f_i\in\mathbb{F}$}
                \If{$i=i^{*}$}
                    \State{$t_i=\theta_d$}
                \Else
                    \State{$t_i=0$}
                \EndIf
            \EndFor
            \State{$\delta_d=\frac{\sum_i(I(z_d,f_i)-t_i)^2}{\theta_d^2({Count}(\mathbb{F})-1)}$}
        \EndFor
        \State{${score}={avg}(1-\delta_d$)}
    \end{algorithmic}
    \textbf{Explicitness:}
    \begin{algorithmic}[1]
        \For{$f_i \in\mathbb{F}$}
            \State{$\mathbb{D}_i=\varnothing$}
            \For{$v_{ij}\in\mathbb{V}_i$}
                \State{$p(v_{ij})=\frac{{Count}(\mathbb{S}_{ij})}{\sum_{j}{Count}(\mathbb{S}_{ij})}$}
                \State{Sample $N_j=N\times p(v_{ij})$ representations $\mathbf{r}^{j}$ from $\mathbb{R}_{ij}$}
                \For{$n=1,2,\ldots,N_j$}
                    \State{$\mathbb{D}_i=\{(\mathbf{r}_{n}^{j},v_{ij})\}\bigcup\mathbb{D}$}
                \EndFor
            \EndFor
            \State{Split $\mathbb{D}_i$ into training set $\mathbb{TR}_i$ and test set $\mathbb{TE}_i$ with proportion (80\%, 20\%)}
            \State{Train an one-versus-rest logistic regress classifier on $\mathbb{TR}_i$}
            \State{Record the ROC area-under-the-curve (AUC) ${auc}_{ij}$ on $\mathbb{TR}_i$ for every $v_{ij}$}
        \EndFor
        \State{${score}={avg}({auc}_{ij})$}
    \end{algorithmic}
\end{algorithm}
\begin{algorithm}
    \caption{Metric of \citet{eastwood2018a}}
    \label{alg:eastwood}
    \begin{algorithmic}[1]
        \For{$f_i \in\mathbb{F}$}
            \State{$\mathbb{D}_i=\varnothing$}
            \For{$v_{ij}\in\mathbb{V}_i$}
                \State{$p(v_{ij})=\frac{{Count}(\mathbb{S}_{ij})}{\sum_{j}{Count}(\mathbb{S}_{ij})}$}
                \State{Sample $N_j=N\times p(v_{ij})$ representations $\mathbf{z}^{j}$ from $\mathbb{R}_{ij}$}
                \For{$n=1,2,\ldots,N_j$}
                    \State{$\mathbb{D}_i=\{(\mathbf{z}_{n}^{j},v_{ij})\}\bigcup\mathbb{D}$}
                \EndFor
            \EndFor
            \State{Split $\mathbb{D}_i$ into training set $\mathbb{TR}_i$ and test set $\mathbb{TE}_i$ with proportion (80\%, 20\%)}
            \State{Train a random forest classifier on $\mathbb{TR}_i$}
            \State{Informativeness score ${inf}_i$ is the accuracy on $\mathbb{TE}_i$}
            \State{$r_{id}$ is the relative importance of dimension $d$ in predicting $v_{ij}$, obtained from the random forest}
            \For{$d=1,2,\ldots,dim_z$}
                \State{$P_d=\frac{r_{id}}{\sum_d r_{id}}$}
            \EndFor
            \State{$H=-\sum_d P_d\log_{dim_z}P_d$}
            \State{$dis_i=1-H$}
        \EndFor
        \State{${score}_{disentanglement}={avg}({dis}_i)$}
        \State{${score}_{informativeness}={avg}({inf}_i)$}
        \For{$d=1,2,\ldots,dim_z$}
            \For{$f_i\in\mathbb{F}$}
                \State{$Q_i=\frac{r_{id}}{\sum_i r_{id}}$}
            \EndFor
            \State{$H=-\sum_i Q_i\log_{{Count}(\mathbb{F})}Q_i$}
            \State{Completeness score $com_d=1-H$}
        \EndFor
        \State{${score}_{completeness}={avg}({com}_d)$}
    \end{algorithmic}
\end{algorithm}

\end{document}